\documentclass[runningheads]{llncs}

\usepackage{eccv}

\usepackage{eccvabbrv}

\usepackage{graphicx}
\usepackage{booktabs}
\usepackage{amssymb}
\usepackage{enumitem}
\usepackage[accsupp]{axessibility}  

\usepackage{hyperref}

\usepackage{orcidlink}

\begin{document}

\title{When the Small-Loss Trick is Not Enough: 
Multi-Label Image Classification with Noisy Labels 
Applied to CCTV Sewer Inspections.} 

\titlerunning{When the Small-Loss Trick is Not Enough}

\author{Keryan Chelouche\orcidlink{0009-0002-1692-1827}\and
Marie Lachaize \and
Marine Bernard \and
Louise Olgiati \and
Remi Cuingnet\orcidlink{0000-0002-9104-2446}}

\authorrunning{K.~Chelouche et al.}

\institute{Veolia Research \& Innovation \\
\email{keryan.chelouche@veolia.com \\ remi.cuingnet@veolia.com}}
\maketitle

\begin{abstract}
The maintenance of sewerage networks, with their millions of kilometers of pipe, heavily relies on efficient Closed-Circuit Television (CCTV) inspections. Many promising approaches based on multi-label image classification have leveraged databases of historical inspection reports to automate these inspections. However, the significant presence of label noise in these databases, although known, has not been addressed.
While extensive research has explored the issue of label noise in single-label classification (SLC), little attention has been paid to label noise in multi-label classification (MLC). To address this, we first adapted three sample selection SLC methods (Co-teaching, CoSELFIE, and DISC) that have proven robust to label noise. Our findings revealed that sample selection based solely on the small-loss trick can handle complex label noise, but it is sub-optimal. Adapting hybrid sample selection methods to noisy MLC appeared to be a more promising approach.
In light of this, we developed a novel method named MHSS (Multi-label Hybrid Sample Selection) based on CoSELFIE. Through an in-depth comparative study, we demonstrated the superior performance of our approach in dealing with both synthetic complex noise and real noise, thus contributing to the ongoing efforts towards effective automation of CCTV sewer pipe inspections.


%
\end{abstract}

\section{Introduction}
\label{sec:intro}

The sanitary sewer system is one of the main public infrastructures playing a vital role in public health and environmental protection. To illustrate the scale of these systems, in the United States, more than 1.2 million kilometers of public sewers serve approximately 240 million Americans~\cite{asce2021} and in China, about 444 million people are served by 511,200 kilometers of sewer pipelines in urban areas~\cite{huang2018current}.
%
%
The maintenance of these infrastructures depends on routine inspections of sewer pipelines using various methods and technologies to detect defects. Closed-Circuit Television (CCTV) is the most widely utilized method~\cite{liu2013state}.
During this procedure, an operator controls a camera-equipped crawler that navigates through the sewer pipeline. The operator visually identifies defects, captures images of the damaged sections, and categorizes each defect using a standardized coding system.

As the volume of collected CCTV inspection data has surged in recent decades, there is a growing interest in leveraging this data to automate the inspection process. Object recognition~\cite{kumar2020deep,wang2021automated,tan2021automatic,dang2022defecttr} and multi-label classification~\cite{haurum2021sewer,hu2023toward,tao2022cafen,xu2024sewer} approaches offer promising solutions. However, these technologies hinge on the availability of high-quality, accurately labeled training data. 

In practice, the reliance on manual inspection processes introduces significant variability in defect recognition~\cite{dirksen2013consistency}. This often manifests as label noise within historical inspection databases. Another significant source of label noise arises from the inherent limitations of the inspection camera. As it cannot capture a complete 360-degree view of the pipe interior, it is not guaranteed that all reported defects are fully visible in their associated images~\cite{haurum2021sewer}. As label noise can severely impede the performance of supervised learning algorithms \cite{algan2020label} due to the memorization effect \cite{arpit2017closer}, addressing this issue is critical.

Given the complexity of sewer pipe defects, where multiple types of damage can occur simultaneously in a single section of pipe, the problem naturally lends itself to a multi-label classification (MLC) framework \cite{haurum2021sewer}. Unlike single-label classification (SLC), which restricts each image to one label, MLC allows each image to be tagged with multiple labels corresponding to different defect types. This is essential to accurately reflect real-world scenarios where multiple defects often co-exist. 
However, while significant research has been conducted on label noise in SLC~\cite{frenay2013classification,song2022survey}, the field of multi-label noise remains relatively under-explored and lacks standardization such as unified evaluation procedure and noise injection strategies~\cite{song2022survey,Burgert_2022}.
Furthermore, naive adaptation of robust SLC methods for noisy MLC proves mostly ineffective, particularly when dealing with realistic and complex noise~\cite{song2022survey,tlmc,Burgert_2022,holistic}.

This paper addresses the challenge of multi-label classification with noisy labels, with a focus on CCTV sewer pipe inspections. The main contributions of this paper are as follows:
\begin{itemize}
    \item   A review of existing noise injection strategies in MLC, along with the development of a standardized method for reporting noise levels in corrupted datasets, facilitating more reliable comparisons across studies.
    \item A comprehensive study of the impact of the type of noise on the performances of sample selection methods for MLC.
    \item A thorough adaptation of three sample selection methods from SLC to MLC, specifically Co-teaching~\cite{han2018coteaching} based on the small-loss trick and two hybrid methods, CoSELFIE~\cite{pmlr-v97-song19b}, and DISC~\cite{li2023learning}, along with an analysis of their performances, providing insights into their effectiveness and limitations.
    \item A new method called MHSS (Multi-label Hybrid Sample Selection), built upon the adapted CoSELFIE and specifically tailored for multi-label improving performances. 
\end{itemize}
To the best of our knowledge, this is the first paper to demonstrate a substantial improvement in performance using a sample selection method for MLC on severely corrupted datasets with complex, mixed noise.

\section{Related Work}
\label{sec:rel_work}

\subsection{Classification with Noisy Labels}
\label{subsec:label_noise}

\subsubsection{Single Label Classification}

Song~\etal~\cite{song2022survey} reported that real-world SLC databases can contain a significant amount of noisy labels, ranging from 8\% to 38.5\%. This noise can hinder model generalization, and traditional regularization techniques like data augmentation, dropout, and batch normalization may not always mitigate its impact. Consequently, there
is a growing body of research focused on developing methods robust to label noise. These methods can be grouped into five categories~\cite{song2022survey}: robust architecture, robust regularization, robust loss, loss adjustment and sample selection.

Robust architecture methods modify the neural network architecture, often by adding a noise adaptation layer~\cite{goldberger2017training}. This aims at estimating the noise distribution in the data, to better handle noisy labels. In addition to these methods, advanced regularization techniques, such as the early-learning regularization (ELR)~\cite{liu2020early}, offer alternatives  for improving robustness. 
%
%
Other approaches either adjust the loss function, like self-adaptive training (SAT)~\cite{huang2022self} that computes the loss with exponentially-moving averaged predictions as ground-truth labels, or use intrinsically robust loss functions such as the generalized cross entropy~(GCE)~\cite{zhang2018generalized}.
A different line of research tackled label noise by selecting samples to train on. This is the case of Co-teaching~\cite{han2018coteaching} and JoCoR~\cite{Wei_2020_CVPR}.
More recent works proposed hybrid methods combining sample selection with other approaches.
For instance, in SELFIE, Song~\etal~\cite{pmlr-v97-song19b} combined it with a loss adjustment by \textit{label refurbishing}. Similarly, in DISC, Li~\etal~\cite{li2023learning} merged sample selection with loss adjustment, robust regularization and robust loss.



\subsubsection{Multi-label classification}  Unlike SLC, few MLC methods have been proposed to address label noise~ \cite{song2022survey,tlmc,ccmn}.
%
%
Xie~\etal~\cite{ccmn} established a comprehensive framework for addressing Class-Conditional Multi-label Noise (CCMN), distinguishing between additive noise (where negative labels become positive) and subtractive noise (where positive labels become negative) types. However, their proposed approach was exclusively tested  within the context of partial multi-label learning~\cite{xie2018partial}, so with additive noise only. 
Ghiassi~\etal~\cite{tlmc} introduced TLCM, a promising method outperforming ASL~\cite{ridnik2021asymmetric} on MS-COCO~\cite{lin2014microsoft}  with up to 60\% symmetric label noise. However, their approach requires both a subset of clean data and an additional set of single-label clean data, which are rarely available.
%
To overcome these limitations, Xia et al. \cite{holistic} proposed the holistic label correction (HLC) method. It  handles multi-label noise by leveraging label dependence to correct noisy labels. Yet, HLC is not model agnostic. It is built upon the GCN-based method ADDGCN \cite{ye2020add}, while GCN based methods have been outperformed by TResNet~\cite{Ridnik_2021_WACV,haurum2021sewer}.
%
Instead, to handle multi-label noise, Burgert~\etal~\cite{Burgert_2022} adapted existing robust SLC methods: SAT~\cite{huang2022self}, ELR~\cite{liu2020early}, and JoCoR~\cite{Wei_2020_CVPR}. Their adaptations of SAT and ELR managed to improve classification performances in scenarios with additive, subtractive and low to moderate mixed noise. Their adaption of the JoCoR sample selection method did not consistently yield better classification performances for subtractive and mixed noise. Notably, 
they reported that additive noise alone had little impact on performances while \emph{mixed} noise, combining both additive and subtractive noise, presented the greatest challenge.
%
%
%

\subsection{Sewers}
\label{subsec:sewers}
In order to automate sewer inspection, some research teams proposed to use state-of-the-art object detection algorithms to detect and identify potential defects in the pipe~\cite{ren2016faster,redmon2016you,liu2016ssd,kumar2020deep}. 
To take into account the specificity of sewer pipes, Tan~\etal~\cite{tan2021automatic} and Dang~\etal~\cite{dang2022defecttr} proposed dedicated approaches with promising results.
Other research groups have suggested more advanced defect localization using instance segmentation~\cite{li2022robust, ma2022automatic} or semantic segmentation \cite{dang2023lightweight,li2024pipetransunet, wang2020unified,zhou2022automatic}. Although localizing a defect is useful, it's not essential for codification and it necessitates the creation of a dedicated dataset.

An alternative approach is to use historical inspection reports to extract labeled images of sewers and treat automation as an MLC problem. Haurum and Moeslund~\cite{haurum2021sewer} compared various approaches with methods specifically designed for sewer inspection 
tasks~\cite{xie2019automatic,chen2018intelligent, hassan2019underground, myrans2019automated,kumar2018automated,meijer2019defect} alongside general MLC models such as TResNet~\cite{Ridnik_2021_WACV}, ML-GCN~\cite{chen2019multi} and KSSNet~\cite{wang2020multi}.
As a result, for images with defects, the best performances were obtained with a TResNetL~\cite{Ridnik_2021_WACV}.

However, CCTV sewer inspection datasets are prone to label noise due to inter-operator variability~\cite{dirksen2013consistency} and limited camera view~\cite{haurum2021sewer}. Different inspectors may indeed label the same defect differently, leading to inconsistencies in the data. Furthermore, inspection cameras cannot capture a complete 360-degree view of the pipe interior, potentially causing reported defects not to be visible in captured images. To the best of our knowledge, this issue of label noise in CCTV sewer inspection datasets, while known, has not been addressed in the literature.

\section{Methods}
\label{sec:methods}



\subsection{Adapted methods}
\label{subsec:adapted}
We adapted three sample selection methods from the robust SLC literature, namely Co-teaching~\cite{han2018coteaching}, CoSELFIE~\cite{pmlr-v97-song19b} and DISC~\cite{li2023learning}, as follows. First, softmax functions were replaced by sigmoids and the cross-entropy loss by the asymmetric loss (ASL)~\cite{ridnik2021asymmetric}.
Then, similarly to Burgert~\etal~\cite{Burgert_2022}, instance-level selections and corrections used in SLC were substituted by their label-level equivalents, thereby allowing partial selection and correction more suited to MLC. 
The following describes in more details Co-teaching~\cite{han2018coteaching} and CoSELFIE~\cite{pmlr-v97-song19b}, on which our method (\cref{subsec:our}) is based.

\subsubsection{Co-teaching}
Co-teaching is a sample selection strategy proposed by Han~\etal~\cite{han2018coteaching}, designed to effectively handle noise in single class label datasets. This approach involves simultaneously training two networks.
In each mini-batch, each network selects samples for its peer network to train on. 
This sample selection is carried out using the \emph{small-loss trick}. It consists in selecting a progressively decreasing proportion of instances with the smallest loss. 

The \emph{small-loss trick} is based on the assumption that noisy labels are \textit{harder} to learn and on the empirical observation that deep neural networks memorize easy instances first, and gradually adapt to hard instances as training epochs become large. This is called the \emph{memorization effect}~\cite{arpit2017closer}.  Thus, it assumes that noisy labels yield larger losses during the first training epochs. 

The progressively decreasing proportion of selected instances $(1-\tau)$ is controlled by a single hyper-parameter $\tau$ called the \textit{forget rate}. 
This rate relates to the dataset's noise rate $\varepsilon$ via a proportionality constant $\alpha$, such that:
\begin{equation}
 \tau = \alpha \cdot \varepsilon
  \label{eq:forget_rate}
\end{equation}

Since $\tau$ depends on the dataset, 
we rather considered $\alpha$ as the hyper-parameter to tune. 
%
Notably, in the original Co-teaching paper~\cite{han2018coteaching},  the optimal results were found with $\alpha \in [1.25, 1.5 ]$ and setting $\alpha$ to one often yielded satisfactory results.


\subsubsection{CoSELFIE} CoSELFIE~\cite{pmlr-v97-song19b} is a \textit{hybrid} method
that incorporates \textit{label refurbishment} into Co-teaching. Hence, this approach also involves the concurrent maintenance of two networks. 

After the selection process where each network categorizes instance labels as either \textit{clean} or \textit{noisy}, instance labels are also divided between \textit{refurbishable} and \textit{non-refurbishable} labels. At epoch $t$,  a network deems 
an instance label $y_i$ as \textit{refurbishable} if its  past predictions $\left[\hat{y}_i^{(t-Q+1)}, \hat{y}_i^{(t-Q+2)}, \cdots,\hat{y}_i^{(t-1)},\hat{y}_i^{t}\right]$ have been consistent across the last $Q$ training epochs. The consistency is measured using entropy $H_i^{(t)}$ scaled between $0$ and $1$:
\begin{equation}
        H_i^{(t)} = \log\left(\frac1K\right)\sum_{k=1}^K p_{i,k}^{(t)}\log p_{i,k}^{(t)}, \quad\text{with} \quad p_{i,k}^{(t)} = \frac1Q\sum_{q=0}^{Q-1} \mathbf{1}_{\hat{y}_i^{(t-q)}=k}
\end{equation}
where $K$ is the number of classes in the SLC settings. Note that, in MLC,  $K=2$ for each label. An instance label $y_i$ with a scale entropy $H_i^{(t)}$ lower than a threshold~$\theta$ is regarded as \textit{refurbishable}. In this case, the most frequently predicted label in past $Q$ epochs becomes the new corrected ground-truth. 
%
Finally its peer network is trained on all but \textit{non-refurbishable} noisy  instance labels.

\subsection{Our method}
\label{subsec:our}
This section describes our method MHSS (Multi-label Sample Selection), which is built upon the MLC adaptation of CoSELFIE (\cref{subsec:adapted}). It is based on two key, yet simple, modifications: one concerning the selection and the other  related to the correction process.

\subsubsection{Class-dependent Noise Rate (CDNR)}
Previous approaches based on the \textit{small-loss trick} used a single noise rate $\varepsilon$ for all classes $c\in\mathcal C$. However, in the MLC setting, assuming a constant noise rate across classes is often unrealistic~\cite{holistic}. Therefore, we modified the forget rate $\tau$ to be class dependent. With $\varepsilon_c$ the noise rate of class $c$, it is then defined as: 
\begin{equation}
    \forall c\in\mathcal C,~ \tau_c := \alpha\cdot\varepsilon_c
\end{equation} 

\subsubsection{Joint Correction Criterion (JCC)}
The rationale behind CoSELFIE is that classification performance can be improved by $(i)$ removing noisy labels through the sample selection step and $(ii)$ augmenting the count of clean labels via the label correction phase (\textit{label refurbishment}). By raising the entropy selection threshold~$\theta$, more corrected labels can be included, allowing for the acceptance of instance labels that show less consistency across prior training epochs. However, this raises the risk of including wrongly corrected labels. 

We claim that a more effective method to relax this acceptance criterion is to utilize a \textit{joint correction criterion} leveraging the dual network architecture by considering an instance as \textit{refurbishable} if at least one of the two networks deems it \textit{refurbishable}.
When a label instance is considered as \textit{refurbishable} by both networks, each network corrects the instance labels for its peer.  On the other hand, when only one considers the instance label as \textit{refurbishable}, the corresponding label correction is applied for both networks.

\subsection{Noise Injection}
\label{subsec:noise_inj}

The standard procedure for assessing  the robustness of methods to label noise consists in injecting a controlled synthetic noise into the training datasets. 
In Single Label Classification (SLC), methods are systematically tested on datasets with symmetric noise and with pair noise~\cite{song2022survey}. However there is no consensus in MLC with label noise~\cite{Burgert_2022}. 
While some studies only employed uniform noise injection~\cite{zhao2021evaluating}, more recent research proposed  
more advanced techniques tailored to the specificities of MLC\cite{Burgert_2022,holistic,tlmc}.
However, these techniques are not standardized and are often unique to each study. 
This variability presents a challenge when comparing results. 
Notably, the reported level of label corruption is not equivalent across different studies, making direct comparisons unfeasible.

For the purposes of this paper, we categorized noise injection methods as \textit{naive} or \textit{complex}. 


\subsubsection{Naive Noise Injection}

Some studies introduced noise uniformly by randomly flipping a proportion of the labels according to a fixed noise rate~\cite{zhao2021evaluating}. However, due to the scarcity of positive labels in MLC, this method predominantly generates additive noise (where negative labels become positive) and minimally produces subtractive noise (where positive labels become negative). This imbalance significantly alters the class distribution and the number of labels per instance. Moreover, it has been demonstrated that additive noise tends to have a less adverse impact on performance compared to subtractive noise \cite{Burgert_2022}  

\subsubsection{Complex Noise Injection}
More recent studies introduced MLC specific noise that maintained certain properties of the clean dataset~\cite{Burgert_2022, holistic}.
%
More specifically, Burgert~\etal\cite{Burgert_2022} injected \textit{mixed noise}: Given a noise rate, for each class, the corresponding proportion of positive labels were flipped and then the same number of negative labels were selected and flipped to positive.
This approach ensures an equal proportion of additive and subtractive noise, while preserving the class distribution of the original dataset. 
However, due to the inherent sparsity of positive labels in most multi-label datasets, the actual number of labels that are flipped will be significantly lower than the specified noise rate parameter.

As for Xia et al. \cite{holistic}, they employed a \textit{class-dependent transition matrix} to swap labels between classes within the same instance as opposed to flipping labels from positive to negative or vice versa. Their method ensures an equal proportion of additive and subtractive noise, while preserving the original dataset's number of labels per instance. However, given the sparsity of positive labels in most multi-label datasets, the majority of these swaps occurs between two negative labels. Consequently, similarly to \textit{mixed noise}~\cite{Burgert_2022}, the actual number of flipped labels is much lower than the noise rate parameter reported.



In this study, we defined $\varepsilon$, the level of noise in a dataset, as the percentage of labels that have been genuinely inverted, regardless of the  noise injection technique employed. Having a relevantly defined noise level is necessary to ensure consistent reporting of results. Furthermore, $\varepsilon$ is used to define the forget rate parameter $\tau$.

\section{Experiments on synthetic noise}
\label{sec:exp}

\subsection{Experimental Setup}
\label{subsec:exp_setup}
To assess the robustness to label noise of the methods described in Section~\ref{sec:methods}, we evaluated their performances on two publicly available datasets, UcMerced~\cite{yang2010bag} and TreeSatAI~\cite{ahlswede2023treesatai}, with both \textit{naive} and \textit{complex} noise injections.

\subsubsection{Datasets} 
The UcMerced Dataset, a land use image collection derived from the USGS National Map Urban Area Imagery, consists of 2100 RGB images. These images are divided into 17 diverse classes, encompassing a variety of subjects from vehicles to natural and urban elements.
The dataset is partitioned into 1700 training images, alongside 184 images for validation and 216 for testing. On average, each image in this dataset is annotated with 3.3 labels, corresponding to a positive label prevalence of approximately 20\%.

The TreeSatAI Dataset is a larger image dataset for forest applications with highly reliable labels, collected in German forest management services. For our study, we utilize the \textit{aerial 3-band} version of the dataset, which encompasses a total of 50,381 images distributed across 15 multi-label \textit{genus} classes. The dataset is divided into 36,269 training images, 9,068 validation images, and 5,044 test images. On average, each image in this dataset is annotated with 1.87 labels, which translates to a positive label prevalence of approximately 12.5\%. Visual examples of both datasets are available in Appendix A.


\subsubsection{Noise Injection}
In this study, we used  both \textit{naive} and \textit{complex} noise injections.
For the \textit{naive} noise injection, we simply flipped a proportion of labels of $\varepsilon$. As for the \textit{complex} noise, the \textit{mixed noise} strategy was adopted (\cref{subsec:noise_inj}). It allowed us to preserve the class distributions of the original datasets. 
To be able to compare our results with those obtained by Burgert~\etal\cite{Burgert_2022}, a noise rate $\varepsilon$ of 20\% for the UcMerced Dataset and 12.5\% for the TreeSatAI Dataset was chosen, which corresponded to a 50\% mixed noise reported in their study.





\subsubsection{Experimental setup}

The same training set-up was used for every experiment. We used a TResNetM~\cite{Ridnik_2021_WACV} pretrained on ImageNet~21k~\cite{ridnik2021imagenetk} as a backbone for every model. The loss function was the asymmetric loss (ASL) with its default hyper-parameters~\cite{ridnik2021asymmetric}. 
Models were trained on input resolution $256 \times 256$, for 30 epochs, using an Adam optimizer~\cite{kingma2014adam} with default parameters and a 1-cycle policy \cite{smith_2018} with a maximum learning rate of $10^{-3}$. The batch size was set to 64 for all experiments, except for DISC, where it was reduced to 32 due to memory constraints. For data augmentation, we used RandAugment \cite{cubuk2020randaugment}. On the UcMerced dataset, each experiment was conducted three times to ensure the robustness and reliability of our findings. However, due to computational costs, experiments on the TreeSatAI dataset were conducted only once.

For evaluation, we followed conventional settings and reported the mean average precision (mAP). 
%
Additionally, to evaluate the label selection process, we monitored the \textit{label precision} (ratio of selected clean labels to selected labels) and the \textit{label recall} (ratio of selected clean labels to clean labels) during training:
\begin{equation}
    \text{\textit{label precision}} := \frac{\text{\# clean \& selected}}{\text{\# selected}},~
    \text{\textit{label recall}} := \frac{\text{\# clean \& selected}}{\text{\# clean}}
\end{equation}
All models were compared to a baseline model: TResNetM~\cite{Ridnik_2021_WACV} with ASL~\cite{ridnik2021asymmetric}.



\subsection{Sample selection with the small-loss trick}
\label{subsec:uni_lim}

\subsubsection{Sample selection on different types of noise} 
The performances of Co-teaching were assessed on the UcMerced dataset with \textit{naive} and \textit{complex} noise.
%
Additionally to the baseline model, Co-teaching was compared to a \textit{cheater model} which simulated an ideal sample selection scenario where the Co-teaching selection would have perfectly excluded all noisy samples and  trained on clean data only. 
We tested different forget rate factors $\alpha$~(\cref{sec:methods}) on the validation dataset: 0.25, 0.5, 1 and 1.5. We selected the value that gave the best mAP.

Results are presented in Table \ref{tab:cot_univmix}. We observed that introducing any type of noise adversely affected performances, with a more pronounced decrease with complex noise. Moreover, while Co-teaching considerably improved performances with naive noise, it fell short in effectively handling complex noise, showing only slight improvements over the baseline. This underscores the necessity of evaluating methods beyond the scope of naive noise injection. 
On the contrary, the \textit{cheater model}'s performances remained closely aligned with performances on the clean dataset for both types of noise. This shows that, while  sample selection approaches could indeed be beneficial for MLC with noisy labels,  the \textit{small-loss trick} alone is not enough to handle complex label noise in MLC.

\begin{table}[h]
\caption{Comparison of Co-teaching performances on the UcMerced Dataset ($\varepsilon = 0.2$).}
\centering
\begin{tabular}{@{}lccc@{}}
\toprule
\textbf{Method} & \textbf{Clean} & \textbf{Naive noise} & \textbf{Complex noise } \\
& &($\alpha = 1.5$) & ($\alpha = 0.25$) \\
\midrule
Baseline & $80.94 \pm 0.97$ & $48.97 \pm 0.57$ &$ 40.13 \pm 0.38$ \\
Co-teaching \cite{han2018coteaching} & $79.88 \pm 0.20$ & $72.77 \pm 0.77$ & $41.90 \pm 1.24$\\
Cheater & - & $79.88 \pm 0.54$ & $78.47 \pm 0.86$ \\
\bottomrule
\end{tabular}
\label{tab:cot_univmix}
\end{table}

\subsubsection{Forget Rate Analysis} To comprehend the failure of the \textit{small-loss trick} in dealing with \textit{complex} noise, we compared the performances of  Co-teaching with different forget rate factors $\alpha$ in the case of \textit{naive} and \textit{complex} noise injection (\cref{tab:cot_fr}).
%
We note a contrasting trend in the performance. Under naive noise injection, the mAP increased with the value of $\alpha$. 
On the contrary, under \textit{complex} noise injection, performance decreased as $\alpha$ increased past 0.25. 


\begin{table}[h]
\caption{ Co-teaching performances with varying $\alpha$ on UcMerced dataset ($\varepsilon = 0.2$) with \textit{naive} and \textit{complex} noise.}
\label{tab:cot_fr}
\centering
\begin{tabular}{l@{\quad}c@{\quad}c@{\quad}c@{\quad}c@{\quad}c}
\toprule
$\mathbf\alpha$& 0 & 0.25 & 0.5 & 1 & 1.5 \\
\midrule
\textit{naive} & $47.51 \pm 0.69$ & $48.62 \pm 0.51$ & $52.47 \pm 0.94$ & $64.64 \pm 1.75$ & $\mathbf{72.77 \pm 0.77}$ \\
\textit{complex} & $39.36 \pm 0.81$ & $\mathbf{41.90 \pm 1.24}$ & $40.35 \pm 0.11$ & $36.96 \pm 0.86$ & $35.05 \pm 0.37$ \\
\bottomrule
\end{tabular}
\end{table}
Intuitively, an $\alpha$ factor of less than one wouldn't fully eliminate noisy labels, even with an optimal selection strategy. 
Conversely, assuming an efficient selection strategy, a greater $\alpha$ would discard a larger proportion of noisy labels. 
However, a significantly higher $\alpha$ (larger than 1) would eventually exhaust the pool of noisy samples, leading to the unintended discard of a substantial number of clean samples. Han~\etal~\cite{han2018coteaching} demonstrated in their research that optimal results are typically achieved with $\alpha$ ranging from 1.25 to 1.5, although a value of 1 consistently yields satisfactory outcomes. In subsequent studies, whether applied to multi-class scenarios or multi-label adaptations, researchers often default to an $\alpha$ value of 1 or do not specify the $\alpha$ value used. The fact that our best results were obtained with an $\alpha$ value considerably less than 1 is then unexpected and warrants further investigation.

\begin{figure}[ht]
\centering
\begin{subfigure}{.5\textwidth}
  \centering
  \includegraphics[width=1\linewidth]{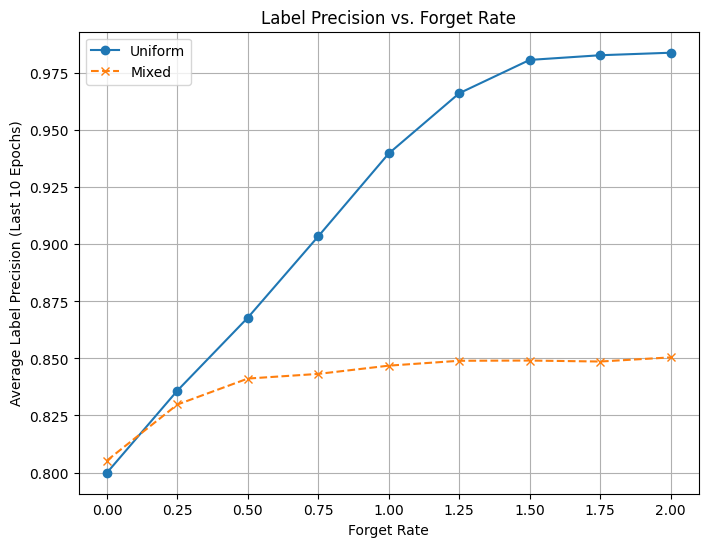}
  \caption{}
  \label{fig:label_precision}
\end{subfigure}%
\begin{subfigure}{.5\textwidth}
  \centering
  \includegraphics[width=1\linewidth]{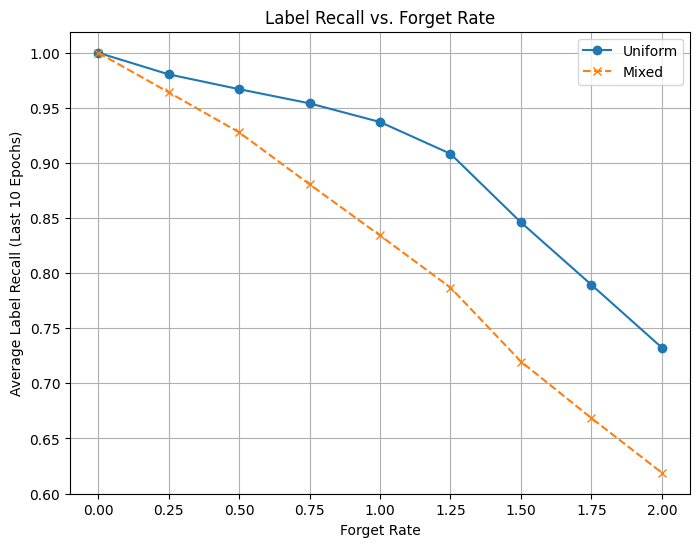}
  \caption{}
  \label{fig:label_recall}
\end{subfigure}
\caption{ Co-teaching average (a) \textit{label precision} and (b) \textit{label recall} over the last 10 epochs versus the forget rate factor $\alpha$ for \textit{naive} uniform and \textit{complex} mixed noise}
\end{figure}

To further analyse the \textit{small-loss trick} selection, we graphed the label recall and precision (as defined in \ref{subsec:exp_setup}) averaged over the final 10 epochs of training in relation to the forget rate factor $\alpha$. In Figure \ref{fig:label_precision}, we note a linear increase in label precision from 0 to 1 in the case of \textit{naive} noise. This increase continues, albeit at a slower pace, from 1 to 1.5, eventually reaching a plateau at a label precision of 0.98. This result aligns with the optimal range of the forget rate factor $\alpha$, previously established in the context of SLC~\cite{han2018coteaching}. However, the situation is markedly different with \textit{complex} noise. In this case, label precision hardly increased past $\alpha = 0.25$, and hit a plateau at $\alpha = 1$ just short of 0.85. This stark contrast highlights the limitations of the \textit{small-loss trick} when dealing with complex noise. As shown by the label recall curve~(\cref{fig:label_recall}), increasing $\alpha$ beyond 0.25 only resulted in training on a reduced dataset without reducing the amount of noise remaining in the dataset. 





\subsection{Methods Comparison}
\label{subsec:meth_comp}
As described in Section~\ref{subsec:our}, our method is an extension of  CoSELFIE, which in turn is based on Co-teaching~(\cref{subsec:adapted}). 
For a more comprehensive view, not only did we compare our method to both, we also included an adaptation of DISC~\cite{li2023learning}. DISC is a state of the art hybrid sample selection not based on the small-loss trick.
We tested different values for the forget rate factor $\alpha$ (\cref{subsec:adapted}) on the validation dataset: 0.25, 0.5 and 1 for both datasets. We selected the parameter values that gave the best mAP. We also tested different values of the entropy threshold ($\theta \in \{0.05, 0.1, 0.15\}$). We selected $\theta$ equal to 0.05 as it gave the best performances, regardless of the method.
mAP achieved on UcMerced and TreeSatAI were reported in Table~\ref{tab:results}. All the  methods showed improvement over baseline.  Our proposed method achieved the best performances on both datasets. It reached  a mAP of 50.03\% and 43.71\% on UcMerced and TreeSatAI respectively. DISC obtained the second best performance on UcMerced with a 48.28\% mAP although it underperformed compared to CoSELFIE on TreeSatAI, scoring 31.07\% versus 35.99\%. Additional results with varying $\varepsilon$ are available in Appendix B.

 

\begin{table}[h]
\centering
\caption{Results (mAP in \%) on datasets injected with mixed noise.}
\label{tab:results}
\begin{tabular}{lccc}
\toprule
\textbf{Method}   & \textbf{UcMerced} & \textbf{TreeSatAI}  \\
& $\varepsilon = 0.2$ & $\varepsilon = 0.125$ \\
\midrule
ASL \cite{ridnik2021asymmetric} & $40.13 \pm 0.38$  & 24.29 \\
Co-teaching \cite{han2018coteaching} & $41.90 \pm 1.24$ & 28.48 \\
CoSELFIE \cite{pmlr-v97-song19b} & $43.17 \pm 1.19$ & \underline{35.99} \\
DISC \cite{li2023learning} & \underline{$48.28 \pm 1.14$} & 31.07 \\
\hline
MHSS (Ours) & $\mathbf{50.03 \pm 2.01}$ & \textbf{43.71} \\
\bottomrule
\end{tabular}
\end{table}

Table~\ref{tab:comparisons} further presents comparative analysis of mAP improvements upon the baseline for various methods from the literature applied to the UcMerced and TreeSatAI datasets. Our proposed method demonstrates consistent superior performance across both datasets. 
Interestingly, our implementation of Co-teaching surpasses JoCoR~\cite{Wei_2020_CVPR}, which is surprising as JoCoR usually outperforms Co-teaching in SLC \cite{Wei_2020_CVPR}. This might be due to the critical role of the $\alpha$ parameter tuning in adapting the small-loss-based sample selection for MLC.


\begin{table}[h]
\centering
\caption{mAP improvements upon baseline on the UcMerced and TreeSatAI datasets. 
Results of the methods denoted with an asterisk (*) are taken from~\cite{Burgert_2022}.}
\label{tab:comparisons}
\begin{tabular}{lcc}
\toprule
\textbf{Method}  & \textbf{UcMerced} & \textbf{TreeSatAI} \\
\midrule
SAT* \cite{huang2022self} & +0 & +4 \\
ELR* \cite{liu2020early} & +1 & +5 \\
JoCoR* \cite{Wei_2020_CVPR} & +0 & +3 \\
\midrule
Co-teaching \cite{han2018coteaching} & +1.8 & +4.2 \\
CoSELFIE \cite{pmlr-v97-song19b} & +3 & \underline{+11.7} \\
DISC \cite{li2023learning} & \underline{+8.1} & +6.8 \\
\midrule
MHSS (Ours) & \textbf{+9.9} & \textbf{+19.4} \\
\bottomrule
\end{tabular}
\end{table}

\subsubsection{Insights into performance improvements} To better understand why our proposed method achieved superior performances, we evaluated the performance of CoSELFIE and our method for different values of $\alpha$. As observed in Table \ref{tab:selfie_fr}, our method and CoSELFIE exhibited similar performances on UcMerced and TreeSatAI for an $\alpha$ of 0.25. However, as  $\alpha$ increased, our method began to substantially outperform CoSELFIE.




\begin{table}[h]
\centering
\caption{Performances of CoSELFIE and our method for varying value of $\alpha$.}
\label{tab:selfie_fr}
\begin{tabular}{l@{\quad}l@{\quad}c@{\quad}c@{\quad}c}
\toprule
 Dataset & Method & $\alpha = 0.25$ & $\alpha = 0.5$ & $\alpha = 1$  \\
\midrule
UcMerced & CoSELFIE & $\mathbf{43.12 \pm 1.21}$ & $41.96 \pm 1.92$ & $41.47 \pm 1.33$\\
($\varepsilon = 0.2$)& MHSS (Ours) & $43.17 \pm 1.19$ & $\mathbf{50.03 \pm 2.01}$ & $47.58 \pm 1.52$\\
\midrule
TreeSatAI & CoSELFIE & 33.21 & 35.70 & 35.99 \\
($\varepsilon = 0.125$)& Ours & 34.31 & 38.38 & \textbf{43.71} \\
\bottomrule
\end{tabular}

\end{table}

\begin{figure}[ht]
\centering
\begin{subfigure}{.5\textwidth}
  \centering
  \includegraphics[width=1\linewidth]{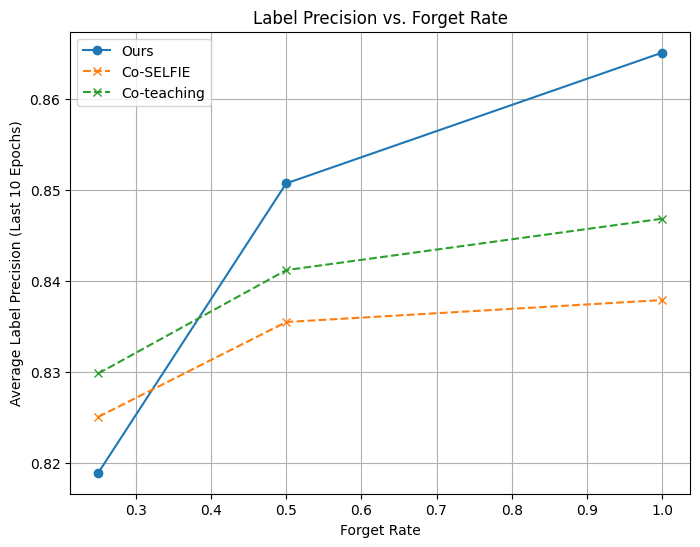}
  \caption{}
  \label{fig:selfie_label_precision}
\end{subfigure}%
\begin{subfigure}{.5\textwidth}
  \centering
  \includegraphics[width=1\linewidth]{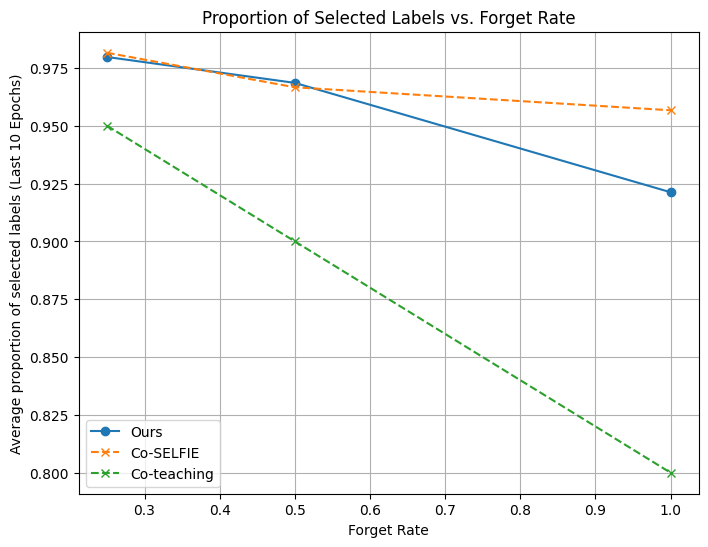}
  \caption{}
  \label{fig:selfie_select}
\end{subfigure}
\caption{ (a) Average \textit{label precision} and (b) average proportion of selected labels  over the final 10 training epochs on UcMerced in relation to the forget rate factor $\alpha$ for Co-teaching, CoSELFIE and our method.}

\label{fig:images}
\end{figure}

To gain a deeper understanding of these results, we graphed in~\cref{fig:selfie_label_precision} the average label precision (\cref{subsec:exp_setup}) over the final 10 training epochs on UcMerced in relation to the forget rate factor $\alpha$ for Co-teaching, CoSELFIE and our method. We also similarly plotted the proportion of selected labels on which training is actually performed in~\cref{fig:selfie_select}. We can observe that, for our method, $\alpha \geq 0.5$ led to a better label precision than for Co-teaching and CoSELFIE. In addition, both our method and CoSELFIE had a very high proportion of selected labels (over 92\%) for all value of $\alpha$. 
As shown by these experiments, performance improvement of our method is related to its capability to better reject or correct noisy labels with a higher $\alpha$.

\subsection{Ablation study}

We performed an ablation study of our method MHSS. Results are presented in Table~\ref{tab:ablation}. We started from the baseline and progressively added the different components of our method. 
\textit{Selection} alone is equivalent to Co-teaching.  \textit{Selection and Correction} corresponds to CoSELFIE. CDNR and JCC are described in details in Section~\ref{subsec:our}. 
Every component individually improved performances on UcMerced, with the biggest impact being the inclusion of either CDNR or JCC to CoSELFIE. However, on TreeSatAI, the addition of CDNR alone to Co-teaching actually decreased performance, same thing with JCC to CoSELFIE. It's only when combined together that those components had a positive impact.  

Our method outperformed every combination of components in both datasets. The addition of \textit{Correction} also always improved performances, highlighting the pertinence of hybrid sample selection methods for MLC. 


\begin{table}[h]
\caption{Ablation study of MHSS}
\centering
\begin{tabular}{|cccc|cc|cc|}
\hline
Selection~ & ~CDNR~ & ~Correction~ & ~JCC~ & \multicolumn{2}{c|}{~UcMerced} & \multicolumn{2}{c|}{~TreeSatAI} \\
& & & &\multicolumn{2}{c|}{($\alpha = 0.5$)}&\multicolumn{2}{c|}{($\alpha = 1$)} \\
\hline
& & & & 40.13 & & 24.29 & \\
\checkmark & & & & 40.35 & +0.22 & 28.48 & +4.19 \\
\checkmark & \checkmark & & & 42.12 & +1.99 & 24.60 & +0.31 \\
\checkmark & & \checkmark & & 41.96 & +1.82 & 35.99 & +11.70 \\
\checkmark & & \checkmark & \checkmark & \underline{48.09} & \underline{+7.96} & 32.10 & +7.81 \\
\checkmark & \checkmark & \checkmark & & 47.21 & +7.08 & \underline{42.62} & \underline{+18.33} \\
\hline
\checkmark & \checkmark & \checkmark & \checkmark & \textbf{50.03} & \textbf{+9.90} & \textbf{43.71} & \textbf{+19.42} \\

\hline
\end{tabular}

\label{tab:ablation}
\end{table}



\section{Experiments on real noise}
\label{sec:exp_advise}
\subsubsection{Dataset}

Our database is composed of 249,093 images from 175,011 sewer inspection reports complying with  European standard EN 13508-2 A1~\cite{nfen13508}. 
We used 70\% for training , 10\% for validation  and 20\% for testing
We ensured that no image from the same inspection report was present in more than one set. 

Such dataset are known to have label noise due to $(i)$ inconsistencies between operators in on defect reporting~\cite{dirksen2013consistency} and $(ii)$ the fact that it is not guaranteed that all reported defects are fully visible in their associated image~\cite{haurum2021sewer}.
To build a clean test set, eight experts annotated a subset of 6,131 images from the test set.  Each image was annotated by three of them and the ground truth labels were estimated from the multiple annotations using the Dawid-Skene approach~\cite{dawid1979maximum}.
To simplify the procedure, we considered a subset of 12 defects\footnote{Defect list:  BAB, BAC, BAG, BAH, BAI, BAJ, BBA, BBB, BBC, BBD, BCA, BCC.} (cf. Appendix~A).
 %
The resulting cleansed test dataset was the one used for evaluation.

\subsubsection{Experimental setup}
The method that achieved the highest performances on the SewerML benchmark~\cite{haurum2021sewer} is a two-stage network. The first stage distinguishes between images with and without defects with the binary classifier from Xie~\etal\cite{xie2019automatic}, while the second stage classifies the type of defect with a TResNetL~\cite{Ridnik_2021_WACV}. Our dataset is only composed of images from inspection reports, so we ignored the first stage and focused solely on the second stage. As a result, we used a TResNetL model~\cite{Ridnik_2021_WACV}  pre-trained on Imagenet~\cite{russakovsky2015imagenet}.
We compared MHSS with the baseline (ASL), Co-teaching and CoSELFIE. We used a forget rate factor $\alpha$ of one. We discarded  DISC since it is based on data augmentation transformations that do not preserve the pipe geometry.
Each model was trained for 30 epochs with a batch size of~32. We used the asymmetric loss (ASL)~\cite{ridnik2021asymmetric} with the default hyperparameters and an Adam optimizer~\cite{kingma2014adam} with a learning rate of $10^{-4}$.

\subsubsection{Results}
Results are presented on~\cref{tab:advise}. The Co-teaching approach 
failed to improve the classification results with an mAP of 55.7\% while the baseline method reached 56.0\% mAP. CoSELFIE and our method (MHSS) achieved a 0.2\% and 0.6\% improvement in mAP respectively. Coupled with our results in Section \ref{subsec:uni_lim}, this finding suggests that real-world noise more closely resembles mixed noise than uniform noise. Consequently, it highlights the importance of developing methods capable of effectively handling mixed noise. 

\begin{table}[h]
\caption{Results on the sewer pipe inspection image dataset.}
\label{tab:advise}
\centering
\begin{tabular}{lc}
\toprule
Method & mAP (\%) \\
\midrule
ASL & 56.0  \\
Co-teaching & 55.7 \\
CoSELFIE & 56.2 \\
\midrule
MHSS (Ours) & \textbf{56.6} \\
\bottomrule
\end{tabular}
\end{table}

\section{Conclusion}
\label{sec:ccl}



This paper addressed label noise in multi-label classification (MLC). We adapted three sample selection single-label classification (SLC) methods - Co-teaching, CoSELFIE, and DISC - to MLC. These methods were evaluated on two datasets, UcMerced and TreeSatAI, with synthetically introduced \textit{complex} label noise. Unlike previous works, we reported noise levels as the true proportions of flipped labels to allow for better comparisons across studies. This evaluation revealed three key findings. First, it stressed out the importance of using \textit{complex} realistic noise over \textit{naive} uniform noise. Secondly, it revealed that the small-loss trick could manage noisy label MLC, provided a suitable forget rate factor $\alpha$ and a properly defined noise level $\varepsilon$. However, these methods were found to be less efficient and were outperformed by hybrid methods.

Based on these insights, we proposed a new method built on the CoSELFIE algorithm tailored for MLC. The proposed approach yielded better sample selections and outperformed others in evaluations using both synthetic \textit{complex} noise and real-world scenarios, showing its potential for the automation of CCTV sewer inspection or other industrial applications.



%
%
\bibliographystyle{splncs04}
\bibliography{main}
\end{document}